\documentclass[12pt]{iopart}

\usepackage{bbm}
\usepackage{physics}
\usepackage[ruled,vlined]{algorithm2e}
\usepackage{float}
\usepackage[font=scriptsize]{subfig}
\usepackage{dirtytalk}
\usepackage{graphicx}
\usepackage{hyperref}
\usepackage{caption}

\begin{document}

\title[Optimizing Quantum Variational Circuits with Deep Reinforcement Learning]{Optimizing Quantum Variational Circuits with Deep Reinforcement Learning}

\author{Owen Lockwood}
\address{Department of Computer Science \\
 Rensselaer Polytechnic Institute, Troy NY, USA}
\ead{lockwo@rpi.edu}
\vspace{10pt}
\begin{indented}
\item[]April 2022
\end{indented}

%
%
%
%
%
\begin{abstract}
Quantum Machine Learning (QML) is considered to be one of the most promising applications of near term quantum devices. However, the optimization of quantum machine learning models presents numerous challenges arising from the imperfections of hardware and the fundamental obstacles in navigating an exponentially scaling Hilbert space. In this work, we evaluate the potential of contemporary methods in deep reinforcement learning to augment gradient based optimization routines in quantum variational circuits. We find that reinforcement learning augmented optimizers consistently outperform gradient descent in noisy environments. All code and pretrained weights are available to replicate the results or deploy the models at: \href{https://github.com/lockwo/rl_qvc_opt}{github.com/lockwo/rl\_qvc\_opt}
.

\end{abstract}

\section{Introduction}

In the last decade the spectre of quantum computing has begun to materialize. Recently, there have been a number of claims of hardware that represents some form of a quantum advantage \cite{arute2019quantum, zhong2020quantum, arrazola2021quantum, wu2021strong}. While it is still debated whether \say{Quantum Supremacy} has been reached \cite{pednault2019leveraging, huang2020classical, pan2021simulating}, there is no doubt that these advances represent substantial improvements in quantum computing hardware. There are a number of challenges when working with these devices, such as noise \cite{steane1998space}, decoherence \cite{pellizzari1995decoherence} and even cosmic rays \cite{mcewen2021resolving}. In addition to these common problems, the supposed \say{killer app} of near term quantum hardware, quantum machine learning, faces additional challenges \cite{mcclean2018barren}. Parallel to the developments of quantum computing in the past decade, deep reinforcement learning (RL) has achieved a number of impressive results. From reaching superhuman performance in games such as Chess \cite{silver2018general}, Poker \cite{brown2019superhuman} and Dota 2 \cite{berner2019dota}, to robotic control \cite{haarnoja2018learning} and chip design \cite{mirhoseini2021graph}.

Quantum Machine Learning (QML) seeks to apply the potential advantages of quantum computing for machine learning problems. Quantum machine learning presents a number of significant (both polynomial and exponential) theoretical speedups \cite{biamonte2017quantum, huang2021provably, liu2021rigorous}. QML algorithms have been developed for supervised learning \cite{schuld2018supervised, havlivcek2019supervised, perez2020data}, unsupervised learning \cite{otterbach2017unsupervised, kerenidis2018q, wiebe2015quantum}, and reinforcement learning \cite{chen2020variational, lockwood2020reinforcement, pmlr-v148-lockwood21a, jerbi2021variational}. A shared problem among many of these techniques is the optimization routine. Gradient \cite{cerezo2020cost} and gradient-free \cite{arrasmith2020effect} optimization techniques decrease in efficacy exponentially as the number of qubits grows. Additionally, independent of these phenomena, the presence of noise also induces exponential difficulties optimization \cite{wang2021noise}. While there has been some work in mitigating the effects of barren plateaus \cite{pesah2020absence, grant2019initialization, larocca2021diagnosing}, it remains a pervasive problem in optimizing QML models.

In this work, we propose a reinforcement learning based approach to the problems of optimizing QML systems. Specifically, we train a deep reinforcement learning agent to minimize the loss of random quantum variational circuits of random sizes with random objectives. There have been several previous applications of reinforcement learning to aid with some of the challenges of optimizing QML systems \cite{sordal2019deep, yao2020noise, yao2020policy, wauters2020reinforcement, khairy2020learning, baum2021experimental}. However, many of these works are limited in their applicable problem space, e.g. to only combinatorial/QAOA \cite{farhi2014quantum} routines or only to certain ansatzs/circuit structures. In this work, we shift towards a more general setup. Specifically, we create an ansatz, depth, qubit number, and cost function agnostic training routine (specified up to a maximum in each of these categories). We find that this optimizer can be used to effectively augment gradient based routines in noisy circuit simulations, increasing the performance across a variety of tasks without increasing the complexity of the circuit sampling. 

\section{Background}

\subsection{Reinforcement Learning}

Deep Reinforcement learning (RL) is one of the three main branches of contemporary deep learning. The goal of RL is to have an agent learn to interact with an environment so as to maximize a reward signal \cite{sutton2018reinforcement}. The framework of RL is often formalized as a Markov Decision Process (MDP) with states $\mathcal{S}$, actions $\mathcal{A}$, and rewards $R$. The objective of this RL optimization problem is $J(\pi) = \max_\pi \sum_{t=0}^H  \left [ \mathbbm{E}_{(s_t, a_t) \sim \pi} r(s_t, a_t) \right ] $ with horizon (the number of timesteps in the environment) $H$, state at time t $s_t$, action at time t $a_t$, and reward function $r$. In other words, the goal is to find the policy, $\pi$, which maximizes the expected return. In this work, we employ entropy maximizing RL algorithms, a state of the art class of algorithms that have shown to be especially robust \cite{eysenbach2021maximum}. Experiments were conducted with other SotA model free algorithm such as Proximal Policy Optimization (PPO) \cite{schulman2017proximal} and Twin Delayed Deep Deterministic Policy Gradient (TD3) \cite{fujimoto2018addressing}, however, we found them to be consistently outperformed by entropy based methods. These entropy maximizing algorithms have a slightly modified objective function, $J(\pi) = max_\pi \sum_{t=0}^H \left [ \mathbbm{E}_{(s_t, a_t) \sim \pi} r(s_t, a_t) + \alpha \mathcal{H}(\pi(\cdot | s_t)) \right ] $ \cite{haarnoja2018soft}, i.e. these algorithms seek to maximize the expected reward and the expected entropy of the policy. At $\alpha = 0$, this is the same as the previous objective. Note that for continuous functions, $\mathcal{H}(\pi(\cdot | s_t)) = - \int_\mathcal{A} \pi(a|s_t) log \pi(a|s_t) da$. 

In order to maximize this objective, we utilize Soft-Actor Critic (SAC) \cite{haarnoja2018app}. SAC is a model free, off-policy, actor critic algorithm. The algorithm is composed of five total neural networks, one policy network, two Q networks and two target Q networks. The Q networks use neural networks to approximate the Q function \cite{watkins1992q}, which is an estimation of the expected reward given a state action pair. In the case of maximum entropy RL this takes the form of $Q_\theta(s_t,a_t) = \mathbbm{E}_\theta \left [r_t + \gamma \; log \int_\mathcal{A} e^{Q(s_{t+1}, a)} da \right ]$. The Q functions are updated via the Soft Mean Squared Bellman Error: \begin{equation*}
\begin{aligned}
    \mathcal{L}_{SMSBE}(\theta) = [ Q_\theta(s_t, a_t) - (r(s_t, a_t) + \gamma (Q_{\theta^\prime}(s_{t+1}, a_{t+1}) \\ - \alpha \; log(\pi(a_{t+1}|s_{t+1}))))  ]^2
\end{aligned}
\end{equation*}The target Q networks (with parameters denoted by $\theta^\prime$) serve to prevent premature numerical overestimation of the Q value and are updated via Polyak averaging \cite{polyak1992acceleration}. The policy network is updated in a similar manner to DDPG \cite{lillicrap2015continuous}, via estimation of the gradient from the Q function: $\nabla_{\phi} \alpha \; log(\pi_\phi(s_t, a_t)) + \nabla_{a_t} \alpha \; log(\pi_\phi(s_t, a_t)) - \nabla_{a_t} Q_\theta(s_t, a_t)$. Additionally, the temperature parameter $\alpha$ can be automatically adjusted. 

\subsection{Quantum Machine Learning}

Quantum machine learning is built upon both advancements in quantum computing and in machine learning. Quantum computing advantages often stem from the ability of quantum computers to represent and operate on information that scales exponentially with the number of qubits. In this work, we focus on the Quantum Variational Circuit (QVC) as the machine learning model \cite{benedetti2019parameterized}. QVCs are a type of quantum circuit with learnable parameters. Any number of QVC setups are possible, in this work we consider arbitrary structure QVCs with the gate set $\left \{CNOT, H, R_x, R_y, R_z \right \}$. Note that this is a universal gateset, meaning any circuit can be represented via these gates. The Pauli rotations gates, $R_x(\theta), R_y(\theta), R_z(\theta)$, rotate around the specified axis $\theta$ radians, $R_\nu (\theta) = e^{-i\frac{\theta}{2}\sigma_\nu}$, where $\nu = X, Y, Z$. The controlled NOT (CNOT) gate is a two qubit gate that can induce entanglement in qubits. The aforementioned $\theta$ are the learnable parameters \cite{mcclean2016theory}. The measurement operator (from which the cost function is calculated) we utilize is the Pauli $\hat{Z}$ operator, or the `computational basis'. 

The gradients of these quantum circuits can be calculated using the parameter shift rule \cite{schuld2019evaluating}. The rotation gates, $R_\alpha(\theta) = e^{-i\frac{\theta}{2}\sigma_\alpha}$, can be differentiated via $\pdv{}{\theta_i} = \frac{f(\theta_i + s) - f(\theta_i - s)}{2 sin(s)}$ given $s \in \mathbbm{R}, s \neq k\pi, k \in \mathbbm{Z}$ \cite{mari2021estimating} where $f(\theta) = \langle 0 | U^\dagger (\theta) \hat{Z} U(\theta) |0 \rangle$ and $U(\theta)$ is composed of these single qubit rotation gates. A common choice for s is $\pi/2$ \cite{bergholm2018pennylane}. Gradients can also be calculated in simulations using adjoint differentiation \cite{plessix2006review, luo2020yao} which requires no parameter perturbations (hence making it substantially faster for classical simulations), but is not feasible on quantum hardware.

\section{Approach}

To work with this QVC optimization problem, we must reconceptualize it into an environment compatible with RL agents. To this end we must consider the how states, actions, and rewards can be represented and numerically fed to the agent's neural network. Here we discuss our approach to each of these. First we have the state space problem, i.e. how do we convert the information from the QVC into an useful format for a RL agent to work with? Additionally, how can we effectively convey information about the structure, inputs, etc. that may vary? In this work we utilize two distinct approaches to convert the QVC information into inputs to the RL agent. Both encoding techniques rely on no simulation/statevector information and are fully compatible with any future or existing hardware. The first approach, which we call \say{feature} encoding, is inspired by the FLIP \cite{sauvage2021flip} algorithm. This encoding takes the QVC and returns a matrix with dimensions $\left [max \: parameters, 8 \right ]$, where each row contains the following information about the parameterized gate: current circuit error, current parameter value, gate type, qubit number, qubit layer, max qubits, max depth and input type. This input is then flatten and processed by a MLP in the RL agent. The second encoding scheme we call \say{block} encoding and is inspired by \cite{fosel2021quantum}. Block encoding takes the QVC and returns a 3D array with dimensions $\left [ max \; qubits, max \; depth, 5 \right ]$. Similar to other RL encoding schemes \cite{silver2018general}, each sheet/plane in this array represents information about a certain feature. The first 3 sheets contain the parameters of the associated Pauli rotation gates (i.e. sheet 1 corresponds with $R_x$ gates, and so on), the fourth represents the input type and the final layer is the current error. This matrix is then fed into a CNN for the agent to process. Prior to training the agent, a maximum number of parameters must be enforced (due to the static size of the weight matrices in the RL neural networks) but can be arbitrarily large. Now we consider the action space. The output of the RL agent is a vector that has the same length as the maximum number of parameters, representing the new value for each possible parameter. Not all parameters would necessarily be used, in cases where the number of parameters are less than the maximum the output is simply clipped to match this size. Finally, we consider the reward function. Due to the brittle nature of RL, constructing an effective reward function can be challenging and important for the agent to succeed \cite{henderson2018deep}. Our reward function is the negative mean squared error between the QVC with the parameters provided by the RL agent and the target value (which is randomly chosen during training). This is not required to be the loss function used in deployment, and we conduct experiments with a variety of cross entropy based losses by simply feeding this loss into the agent as the current error. 

With the basics of the RL agent established, we can detail the training setup. At the beginning of a training routine, the maximum qubits, depth, and optimization timesteps are specified. Implicit within these is the maximum number of parameters (max qubits times max depth). The values used in this work (and provided in the pretrained examples) are 20, 20, and 150, respectively. This limits the number of parameters to be at most 400. At the beginning of each environment iteration a circularly entangled circuit is created with random Pauli rotation gates, random depth and a random number of qubits. An input type (either ground state or equal superposition state) is then selected. Finally a readout cost function is chosen. The possible functions are a product of the Z readout values on each qubit, the Z readout on the first qubit, and the sum of Z readout values on each qubit. A target value is then selected and the agent learns to maximize the negative mean squared error between the target value and the output of QVC. 

The agents were then trained using 2 RTX 6000 24 GB GPUs with 12 vCPUs. With this, the total training time was approximately 150 hours. The evaluation time was done on the same hardware and took approximately 70 hours (largely dominated by some slower noisy simulations). The circuit simulation environment code was generated with TensorFlow-Quantum \cite{broughton2020tensorflow} and Stable Baselines 3 \cite{stable-baselines3} was used as the RL package.

While the agent is trained to directly optimize the circuit, given the scope and complexity of the environment (with no restrictions on the hyperparameters of the circuit), the agent is not meant to be used in this capacity. Rather, the agent is meant to augment traditional approaches by providing an alternative set of parameters at each optimization iteration, as outlined in the following algorithm. Essentially, each optimization step is done by taking the parameters that minimize the loss the most out of the two RL agents and gradient descent as shown in the algorithm below. 

\begin{algorithm}[h]
 
 \While{$iteration < max \; iterations$}{
    $L$ = $\mathcal{L} \left ( f \left ( \theta \right )\right )$ \\
    $\theta_g$ = $\theta - \nabla L$ \\
    $\theta_{MLP} = SAC_{MLP} \left ( L \right )$ \\
    $\theta_{CNN} = SAC_{CNN} \left ( L \right )$ \\
    $L_g$ = $\mathcal{L} \left ( f \left ( \theta_g \right )\right )$ \\
    $L_{MLP}$ = $\mathcal{L} \left ( f \left ( \theta_{MLP} \right )\right )$ \\
    $L_{CNN}$ = $\mathcal{L} \left ( f \left ( \theta_{CNN} \right )\right )$ \\
    $\theta = \min_L \left \{ \theta_g, \theta_{MLP}, \theta_{CNN} \right \}$ \\
    
}
 return $\theta$
 \caption{Augmented Algorithm}
\end{algorithm}

\section{Results}

\begin{figure}[h]
  \centering
  \subfloat[Binary Classification of Circles]{\includegraphics[width=0.33\textwidth]{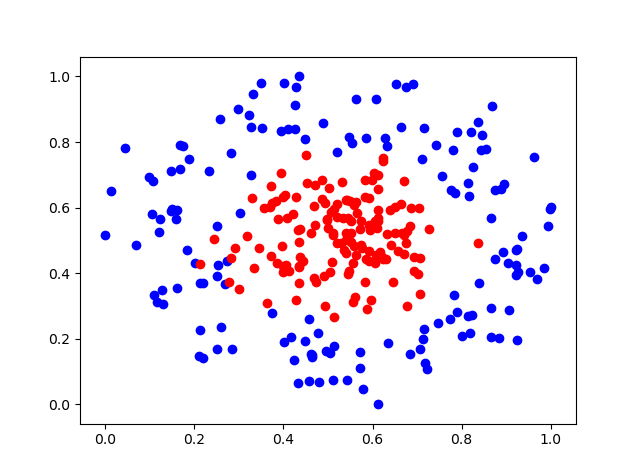}\label{fig:circ}}
  \hfill
  \subfloat[Binary Classification of Moons]{\includegraphics[width=0.33\textwidth]{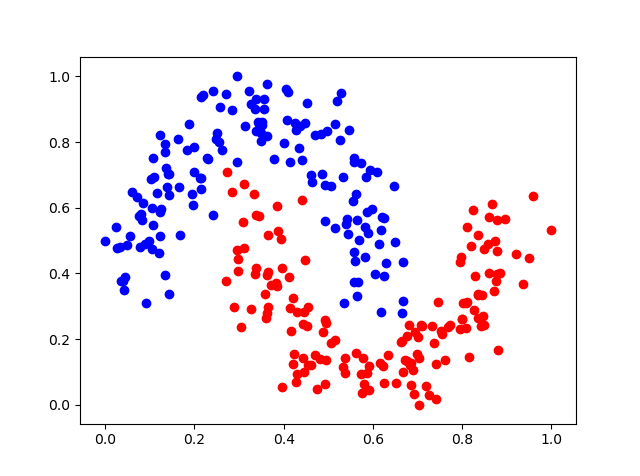}\label{fig:moon}}
  \hfill
  \subfloat[Multi-Class Classification of Blobs]{\includegraphics[width=0.33\textwidth]{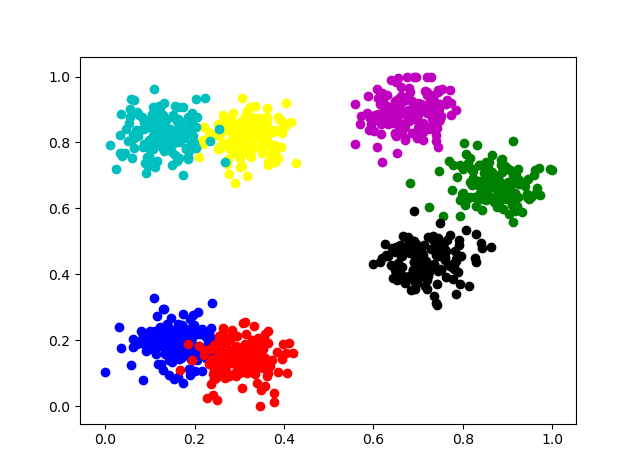}\label{fig:blob}}
  \caption{Classical Data Classification Problems}
  \label{fig:class}
\end{figure}

We evaluate our RL techniques on six different problems, four classical and two quantum. The four classical problems include two binary classification problems, one multi-class classification and one regression (the Boston Housing Dataset). The classification problems can be visualized in Figure \ref{fig:class}. The two binary classification problems utilize 2 qubits, blobs uses 7 qubits, and the regression problem makes use of 13 qubits. The two quantum problems are the optimization routines of the Variational Quantum Eigensolver (VQE) \cite{peruzzo2014variational} and the Quantum Approximate Optimization Algorithm (QAOA) \cite{farhi2014quantum}. For VQE and QAOA we consider 3 different problems sizes of 5, 10, and 20 qubits. Note that the other hyperparameters for VQE and QAOA are constant. The QAOA problem is MAX-CUT and the random graph has regularity two and $p = 10$. The VQE problem generates random Hamiltonians that are decomposed into 10 Pauli sums and utilizes 5 layers of a hardware efficient ansatz \cite{kandala2017hardware}. The results are presented in three tables, each experiment was repeated 3 times (with different random initializations) and the $\pm$ indicates one standard deviation. Note that one cannot compare the numerical values across tables as the problems are randomly generated for each table. Each column represents an optimization technique to the same problem. 
The SAC MLP and SAC CNN columns represent using just the specified RL agent for every optimization step (MLP corresponds with feature encoding and CNN with block). These results of these optimizers used solo can be found in Appendix A. 
For all experiments the gradient descent optimizer is Adam \cite{kingma2014adam} 
. In Table \ref{tab:exp1}, we show the results for simulations with zero noise (shot or depolarizing). The gradient descent makes use of adjoint differentiation to enable larger simulations. Table \ref{tab:exp2} shows the results for simulations with only shot noise. This table is shorter as the larger simulations are less feasible to conduct while using parameter shift differentiation techniques. The noisy simulation results are shown in Table \ref{tab:exp3}. In addition to shot noise, these circuits are simulated with depolarizing noise, modifying the density operator via $\rho \rightarrow \left ( 1 - p \right ) \rho + \frac{p}{4^n - 1} \sum_i P_i \rho P_i $ where $p$ is the probability and $P_i$ are the Pauli gates.  In these experiments $p = 0.075$. The results tend to show consistent advantages for the augmented optimizer in the \say{real world} regimes (i.e. at least noise from expectation approximation) and inconsistent relative performance in completely noiseless simulations (achieving the best results in approximately 1/3 of the experiments).


\begin{table}
\centering
\begin{tabular}{|l|c|c|}
\hline
Evaluation & Gradient Descent & Augmented \\
\hline
Circle Train Loss & $ \mathbf{0.5834 \pm 2 * 10^{-6}} $ & $ 0.5834 \pm 5 * 10^{-6}$ \\
\hline
Circle Validation Loss & $ 0.5761 \pm 0.0001$ & $ \mathbf{0.57048 \pm 0.0017} $ \\
\hline
Moons Train Loss & $ 0.354218 \pm 0.000379 $ & $ \mathbf{0.3538 \pm 3 * 10^{-5} }$ \\
\hline
Moons Validation Loss & $ 0.36714 \pm 0.0042747 $ & $ \mathbf{0.3586 \pm 0.00143}$ \\
\hline
Blobs Train Loss  & $ \mathbf{ 1.5677 \pm 0.03122} $ & $ 1.578 \pm 0.01315 $ \\
\hline
Blobs Validation Loss & $ \mathbf{1.56367 \pm 0.02969} $ & $ 1.5798 \pm 0.0155 $ \\
\hline
Regression Train Loss& $ \mathbf{0.0254 \pm 0.00502} $ & $ 0.0719 \pm 0.0182 $ \\
\hline
Regression Validation & $ \mathbf{0.0291 \pm 0.00513} $ & $0.0747 \pm 0.0131 $ \\ 
Loss  & & \\
\hline
5 Qubit QAOA Cost & $ -1.50 \pm 8 * 10^{-7} $ & $ \mathbf{ -1.50 \pm 9 * 10^{-7}} $ \\
\hline
10 Qubit QAOA Cost  & $\mathbf{ -2.9998 \pm 0.000205} $ & $ -2.9997 \pm 0.000262 $ \\
\hline
20 Qubit QAOA Cost & $\mathbf{ -7.8989 \pm 0.0877} $ & $ -7.667 \pm 0.461 $ \\
\hline
5 Qubit VQE Cost &  $ -2.51396 \pm 0.1517 $ & $ \mathbf{-2.5595 \pm 0.1471} $ \\
\hline
10 Qubit VQE Cost & $ \mathbf{-1.2362 \pm 0.0109} $ & $ -1.2345 \pm 0.0195 $ \\
\hline
20 Qubit VQE Cost & $ -0.05 \pm 0.000838 $ & $\mathbf{ -0.05 \pm 0.000827} $ \\
\hline
\end{tabular} \caption{Noiseless} \label{tab:exp1}
\end{table}  

\begin{table}
\centering
\begin{tabular}{|l|c|c|}
\hline
Evaluation &  Gradient Descent & Augmented \\
\hline
Circle Train Loss  & $\mathbf{0.5674 \pm 0.00032}$ & $0.5676 \pm 0.0003$ \\
\hline
Circle Validation Loss & $0.6087 \pm 0.001076$ & $\mathbf{0.605 \pm 0.0005565}$ \\
\hline
Moons Train Loss  & $0.344 \pm 0.0002$ & $\mathbf{0.3439 \pm 0.00042}$ \\
\hline
Moons Validation Loss & $0.3844 \pm 0.000683$ & $\mathbf{0.3835 \pm 0.0008}$ \\
\hline
5 Qubit QAOA Cost  & $-1.561 \pm 0.0061$ & $\mathbf{-1.5627 \pm 0.0034}$ \\
\hline
10 Qubit QAOA Cost & $\mathbf{-4.8 \pm 0.2382}$ & $-4.687 \pm 0.2756$ \\
\hline
5 Qubit VQE Cost  & $\mathbf{-3.575 \pm 0.198}$ & $-3.5626 \pm 0.076$ \\
\hline
10 Qubit VQE Cost  & $-1.0922 \pm 0.0813$ & $\mathbf{-1.1215 \pm 0.0599}$ \\
\hline
\end{tabular} \caption{Only shot noise}\label{tab:exp2}
\end{table}  

\begin{table}
\centering
\begin{tabular}{|l|c|c|}
\hline
Evaluation & Gradient Descent & Augmented \\
\hline
Circle Train Loss & $0.6739 \pm 0.000303$ & $\mathbf{0.6724 \pm 0.00012}$ \\
\hline
Circle Validation Loss & $0.656 \pm 0.000278$ & $\mathbf{0.6547 \pm 0.00106}$ \\
\hline
Moons Train Loss  & $0.644 \pm 0.005$ & $\mathbf{0.6347 \pm 0.00307}$ \\
\hline
Moons Validation Loss & $0.6406 \pm 0.0058$ & $\mathbf{0.631 \pm 0.0043}$ \\
\hline
5 Qubit QAOA Cost & $-1.52 \pm 0.00998$ & $\mathbf{-1.5773 \pm 0.01062}$ \\
\hline
10 Qubit QAOA Cost & $-4.805 \pm 0.683$ & $\mathbf{-5.2257 \pm 0.0889}$ \\
\hline
5 Qubit VQE Cost  & $ -0.1634 \pm 0.0297 $ & $\mathbf{-0.1716 \pm 0.0242} $ \\
\hline
10 Qubit VQE Cost & $-1.0683 \pm 0.06495$ & $\mathbf{-1.10984 \pm 0.0272}$ \\
\hline
\end{tabular} \caption{Shot and depolarizing noise}\label{tab:exp3}
\end{table}  

\begin{figure}
    \centering
    \includegraphics[scale=0.5]{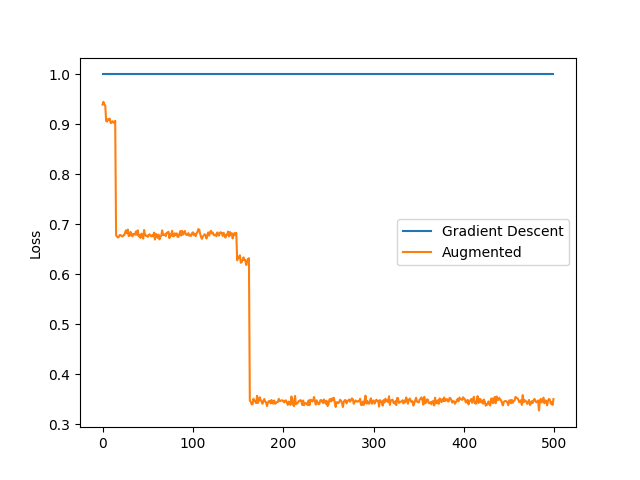}
    \caption{Loss vs. Iterations of 6 Qubit System with Global Cost Function}
    \label{fig:bp}
\end{figure}

\begin{figure}
    \centering
    \includegraphics[scale=0.5]{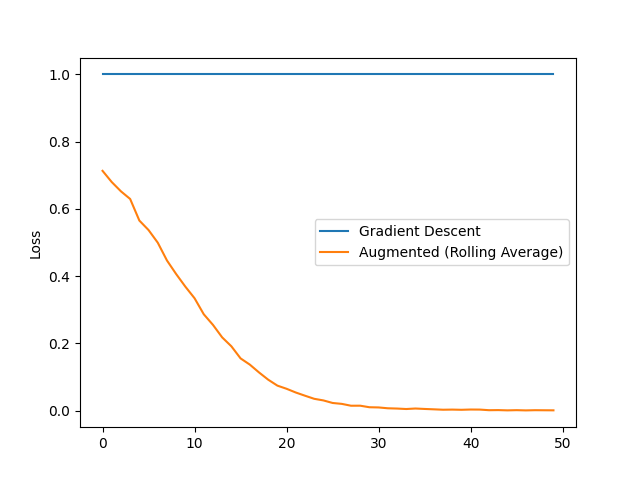}
    \caption{Loss vs. Iterations of 6 Qubit System with Global Cost Function With Rolling Gradients}
    \label{fig:bp1}
\end{figure}

We also briefly present the present the potential of this augmented approach to aid with the barren plateaus problem. We consider a toy example of two layers of $R_X, R_Y$ rotations on 6 qubits. Traditional gradient descent approaches are unable to succeed (even with 10000 shots to reduce shot noise). However, we present two approaches that may help to alleviate this problem. First is the exact same algorithm as above, these results can be seen in Figure \ref{fig:bp}. We are also able to achieve better results in this case by keeping slightly more history and instead of performing gradient descent on $\theta_t$ we also perform gradient descent on $\theta_{t-1}$ and simply add this into the greedy selection of minima for the next $\theta$. This is able to achieve the results shown in Figure \ref{fig:bp1}.

\section{Discussion}

Here it is important to highlight and clarify a potential confusion. It may not be clear as to why the augmented optimizer is not upper bounded by the other optimizers and does not always perform as well. This is primarily because it does not necessarily explore the same region of the cost landscape. At every iteration, the augmented optimizer takes the biggest step in the negative direction, hence it does not follow the same trajectories as the other optimizers. As the RL optimizers' predictions are dependent on the current parameters, these differing steps result in different inputs to the agent and therefore different output. This is likely why the gradient descent approach out performs the augmented optimizer in completely noise free situation, i.e. the greedy augmented approach leads to differing local minima. Secondly, although the RL agents are trained as solo optimizers (without any augmentation), they seem to consistently be the worst performing. This is likely because the evaluations are so different from the training. Given the variety of hyperparameters of current QVC systems (ansatz, cost function, etc.) not only is it impossible to directly train on all combinations, but the goal is to be somewhat generalizable to situations not directly in the training data. Thus, excellent performance on a limited set of tasks is traded for worse performance on a larger set of problems. 

\subsection{Future Work}

There are a number of potential future directions for this line of work. A clear extension is to expand training and validation for larger systems, up to and beyond 30 qubit simulations which would require substantially more computational power. Another important step would be to verify and experiment with these optimizers on actual quantum hardware. Additionally, this approach can be compared and combined with other gradient (and non-gradient) based optimizers to provide further insight into this approach. Finally, the potential to alleviate barren plateaus seems substantial but is limited to a toy problem. Expanding on this analysis is important and potentially very impactful.

\section{Conclusion}

In this work, we presented and experimented with an approach to train and evaluate deep reinforcement learning to optimizer quantum variational circuits. These agents have the potential to take advantage of recent advancements in deep learning and reinforcement learning to aid with the difficult task of quantum circuit optimization. We trained (and provided) models on large systems of up to 20 qubit, 400 parameter quantum variational circuits. We analyzed these agents capabilities to augment existing gradient based approaches to optimization on a variety of quantum machine learning tasks of various sizes and various levels of noise. The empirical findings suggest the augmentation can help even in the absence of noise, and consistently helps in the presence of shot or depolarizing noise. Our work is indicative that contemporary deep learning research can help to alleviate some challenges of working with and optimizing on current and near term quantum hardware.



\newpage

\section*{References}
\bibliographystyle{iopart-num}
\bibliography{references}

\newpage

\appendix

\section{Additional Results}

\begin{table}[H]
\centering
\begin{tabular}{|l|c|c|}
\hline
Evaluation & SAC MLP & SAC CNN  \\
\hline
Circle Train Loss & $ 0.588 \pm 0.00115 $ & $ 0.5993 \pm 0.0056 $  \\
\hline
Circle Validation Loss & $ 0.5657 \pm 0.00148 $ & $ 0.576 \pm 0.01$ \\
\hline
Moons Train Loss & $ 0.4148 \pm 0.0201 $ & $ 0.4266 \pm 0.02148 $ \\
\hline
Moons Validation Loss & $ 0.39574 \pm 0.01771 $ & $ 0.4098 \pm 0.03125 $ \\
\hline
Blobs Train Loss & $ 1.854 \pm 0.00951 $ & $ 1.856 \pm 0.007386 $  \\
\hline
Blobs Validation Loss & $ 1.852 \pm 0.00759 $ & $ 1.859 \pm 0.00538 $ \\
\hline
Regression Train Loss & $ 0.1349 \pm 0.01003 $ & $ 0.1428 \pm 0.00777 $  \\
\hline
Regression Validation & $ 0.1526 \pm 0.00251 $ & $ 0.1572 \pm 0.00395$\\ 
Loss  & &  \\
\hline
5 Qubit QAOA Cost & $ -1.312 \pm 0.09788 $ & $ -1.3387 \pm 0.0979$  \\
\hline
10 Qubit QAOA Cost & $ -2.4763 \pm 0.3418 $ & $ -2.237 \pm 0.319 $ \\
\hline
20 Qubit QAOA Cost & $ -5.3057 \pm 0.194 $ & $ -4.8765 \pm 0.0498$ \\
\hline
5 Qubit VQE Cost & $ -1.604 \pm 0.00368 $ & $ -0.6838 \pm 0.0863 $  \\
\hline
10 Qubit VQE Cost & $ -0.1667 \pm 0.031 $ & $ -0.1598 \pm 0.00192$  \\
\hline
20 Qubit VQE Cost & $ -0.00457 \pm 0.0004 $ & $ -0.00479 \pm 0.0007 $  \\
\hline
\end{tabular} \caption{Noiseless} \label{tab:exp1_app}
\end{table}  

\begin{table}[h]
\centering
\begin{tabular}{|l|c|c|}
\hline
Evaluation & SAC MLP & SAC CNN  \\
\hline
Circle Train Loss & $0.5743 \pm 0.00063$ & $0.5876 \pm 0.0103$  \\
\hline
Circle Validation Loss & $0.6094 \pm 0.0023$ & $0.6138 \pm 0.00868$  \\
\hline
Moons Train Loss & $0.4401 \pm 0.0159$ & $0.4172 \pm 0.0229$ \\
\hline
Moons Validation Loss & $0.4563 \pm 0.008$ & $0.4493 \pm 0.0219$ \\
\hline
5 Qubit QAOA Cost & $-1.293 \pm 0.0451$ & $-1.2663 \pm 0.0404$ \\
\hline
10 Qubit QAOA Cost & $-2.84 \pm 0.4335$ & $-2.634 \pm 0.3651$ \\
\hline
5 Qubit VQE Cost & $-0.9572 \pm 0.2028$ & $-0.8533 \pm 0.078$  \\
\hline
10 Qubit VQE Cost & $-0.209 \pm 0.00234$ & $-0.226 \pm 0.0293$ \\
\hline
\end{tabular} \caption{Only shot noise}\label{tab:exp2_app}
\end{table}  

\begin{table}[h]
\centering
\begin{tabular}{|l|c|c|}
\hline
Evaluation & SAC MLP & SAC CNN  \\
\hline
Circle Train Loss & $0.676 \pm 0.000288$ & $0.681 \pm 0.00131$  \\
\hline
Circle Validation Loss & $0.6571 \pm 0.00022$ & $0.6679 \pm 0.0019$ \\
\hline
Moons Train Loss & $0.6599 \pm 0.00176$ & $0.6657 \pm 0.00204$ \\
\hline
Moons Validation Loss & $0.6564 \pm 0.000924$ & $0.6645 \pm 0.00253$ \\
\hline
5 Qubit QAOA Cost & $-1.364 \pm 0.0936$ & $-1.368 \pm 0.095$ \\
\hline
10 Qubit QAOA Cost & $-3.1087 \pm 0.4198$ & $-2.7073 \pm 0.811$ \\
\hline
5 Qubit VQE Cost & $ -0.1751 \pm 0.0226 $ & $-0.1889 \pm 0.0267$ \\
\hline
10 Qubit VQE Cost & $-0.829 \pm 0.00707$ & $-0.7524 \pm 0.02768$ \\
\hline
\end{tabular} \caption{Shot and depolarizing noise}\label{tab:exp3_app}
\end{table}

\end{document}